\documentclass{IEEEtran}
\usepackage{cite}
\usepackage{amsmath,amssymb,amsfonts}
\usepackage{algorithmic}
\usepackage{graphicx}
\usepackage{textcomp}
\usepackage{color}
\usepackage{statmath}
\usepackage{xcolor,colortbl}
\usepackage{booktabs}
\usepackage{multirow}

\begin{document}

\title{Multi-Feature Fusion and Compressed Bi-LSTM for Memory-Efficient Heartbeat Classification on Wearable Devices}

\author{{Reza Nikandish, \IEEEmembership{Senior Member, IEEE}, Jiayu He, and Benyamin Haghi}
\thanks{R. Nikandish and J. He are with the School of Electrical and Electronic Engineering, University College Dublin, Dublin D04 V1W8, Ireland (e-mail: reza.nikandish@ieee.org).}
\thanks{B. Haghi is with the School of Electrical and Electronic Engineering, California Institute of Technology, Pasadena, CA, USA.}
}

\maketitle 

\begin{abstract}
In this article, we present memory-efficient electrocardiogram (ECG) based heartbeat classification using multi-feature fusion and compressed bidirectional long short-term memory (Bi-LSTM). The dataset comprises five original classes from the MIT-BIH Arrhythmia Database. Discrete wavelet transform and dual moving average windows are used to reduce noise and artifacts in the raw ECG signal, and extract the main points (PQRST) of the ECG waveform. A multi-feature fusion approach is proposed by using \textit{time intervals} and \textit{under-the-curve areas}, to enhance accuracy and robustness.%This method can improve the classification accuracy for the challenging RBBB and LBBB heartbeat classes, by 52.9 and 17.4 percentage point, respectively. 
Using a Bi-LSTM network, rather than a conventional LSTM network, resulted in higher accuracy with a 28\% reduction in the network parameters. Multiple neural network models with varying parameter sizes, including tiny (84\,k), small (150\,k), medium (478\,k), and large (1.25\,M) models, are developed to achieve high accuracy \textit{across all classes}. Overall accuracy is 96.1\% for the large model and 94.7\% for the tiny model. F1 score across all classes is better than 89.1\% for the large model and 85.1\% for the tiny model. The models compressed using post-training quantization techniques can achieve state-of-the-art performance. The compressed large model with 8-bit integer quantization (INT8) features an accuracy of 88.4\% with 1.3\,MB memory. The compressed tiny model with dynamic range quantization (DRQ) can achieve 94.6\% accuracy with only 139\,kB memory.
\end{abstract}

\begin{IEEEkeywords}
Arrhythmia, electrocardiogram (ECG), long short-term memory (LSTM), model compression, post-training quantization, recurrent neural network (RNN), wearable devices.  
\end{IEEEkeywords}

\section{Introduction}

\IEEEPARstart{AI}{} is revolutionizing healthcare, driven by advancements in high-performance computing systems, the availability of massive datasets, and the progress of efficient deep neural networks (DNNs). Deep learning is widely used in a variety of medical applications, including the processing of medical images via convolutional neural networks (CNNs), feature extraction from physiological signals sampled as time-series data using recurrent neural networks (RNNs), and the processing of electronic health records (EHRs) \cite{esteva2019}. Cardiovascular disease (CVD) remains the leading cause of death globally, with deaths rising from 12.1 million in 1990 to 20.5 million in 2021, according to the World Heart Federation (WHF) \cite{WHF2023}. The electrocardiogram (ECG) is the most commonly utilized physiological signal for detecting CVDs \cite{Schl17}. Recently, deep learning has shown promising results in the processing and classification of ECG signals for CVD diagnosis \cite{minchole2019, kachuee18, oh2024, khan21, rajkumar19, petmezas21, obeidat21, haghi23, yan2021, Mahfuz21}.

Recent research has extensively leveraged deep learning \cite{Lecun15} to decipher intricate patterns in ECG data and achieve highly accurate classification. The early progresses used multilayer perceptron (MLP) \cite{mar-2011} and CNN for heartbeat classification \cite{Kiranyaz-2015, Hannun19, rajkumar19, Acharya17, kachuee18, yan2021, Mahfuz21}. In \cite{Hannun19}, arrhythmia classification across 12 classes of ECG data, recorded from 53,549 patients, achieved F1 score of 84\% using a deep CNN with 43\,M parameters.
Using a CNN with 153\,k parameters, \cite{rajkumar19} achieved an accuracy of 93.6\% for arrhythmia classification within the MIT-BIH dataset. \cite{Acharya17} reached a 94\% accuracy on AAMI classes using a CNN with 20\,k parameters. \cite{kachuee18} employed a deep residual CNN with 99\,k parameters and achieved 93\% accuracy on AAMI classes \cite{AAMI98}. In contrast, \cite{yan2021} developed a CNN with 197\,k parameters but achieved a lower accuracy of 92\% on AAMI classes. The highest classification accuracy of 99.90\% on AAMI classes was achieved by \cite{Mahfuz21} using a large model with 4\,M parameters. However, most of these models demand high computational power, making them challenging to implement for real-time classification on resource-constrained wearable devices \cite{MSh18, Wei2020}. Using a hardware-algorithm co-design approach, \cite{haghi23} implemented a lightweight CNN with analog computing circuits, achieving 95\% accuracy on four AAMI classes with 336 parameters. 

%RNNs which are specially developed for time-series data can offer a promising avenue to enhance ECG classification. 
RNNs can concurrently leverage spatial and temporal features extracted from ECG data to improve accuracy and reduce the size of the network. \cite{obeidat21} proposed a hybrid CNN-LSTM model with 64\,k parameters and achieved an accuracy of 98.2\%. Similarly, \cite{petmezas21} developed a hybrid CNN-LSTM model achieving an accuracy of 97.87\%. A bidirectional LSTM (Bi-LSTM) combined with a CNN and residue connections was presented in \cite{xu-2020}, to achieve 95.9\% accuracy on five AAMI classes. In \cite{Qiao-2020}, an accuracy of 99.3\% on five classes of MIT-BIH dataset was presented using a hybrid deep learning model comprising extreme learning machine (ELM), local receptive field (LRF), and Bi-LSTM. In \cite{Saadatnejad-2020}, an algorithm comprising wavelet transform and multiple LSTM networks was proposed for ECG classification. It achieved an accuracy of 99.2\% on seven classes based on AAMI standard. The algorithm met timing requirements of typical wearable hardware, but the memory size of the model was not reported.

%A common problem in many of these works is the reliance on overall accuracy as the main performance metric, while it is essential to ensure high accuracy \textit{across all classes}. Some of these classes may indicate critical cardiovascular conditions, and their accurate detection can potentially save lives.

Recently, the development of transformer-based models has revolutionized various facets of machine learning. Transformers have become popular for natural language processing and image processing tasks due to their self-attention and parallel processing features \cite{Vaswani-2017, Dosovitskiy-2020
}. Nevertheless, their role in time-series data is still subject to further exploration. Interestingly, very recent developments show that improved RNN architectures, such as minLSTM \cite{Feng-2024} and xLSTM \cite{Beck-2024}, can circumvent the scalability limitations of transformers and achieve superior performance for sequence data.

%Recently, the development of transformer-based models has revolutionized various facets of machine learning. Transformers have become popular for natural language processing and image processing tasks due to their self-attention and parallel processing features [Ref]. Nevertheless, their role in time-series data is still subject to further exploration. In contrast to transformers, recurrent neural networks (RNNs), especially bidirectional long-short term memory (Bi-LSTM) networks, are robust in capturing sequential dependencies prevalent in time series or ECG signal classification tasks. In the face of increasing attention for transformers, Bi-LSTMs are vindicated in applications requiring quick response to sequences of data as they can read sequence information both before and after the present sequence. Furthermore, Bi-LSTMs have demonstrated their effectiveness over transformers in performance and resource utilization on ECG signal abnormality detection. This shows the applicability of Bi-LSTMs in portable healthcare devices whthat are sensitive to power usage.

%Ref: A. Katrompas, T. Ntakouris, and V. Metsis, "Recurrence and self-attention vs the transformer for time-series classification: a comparative study," in Proc. Int. Conf. Artif. Intell. Med., Cham, Switzerland, Jun. 2022, pp. 99–109, Springer International Publishing.

Most of the past developments have focused on achieving a high accuracy, overlooking the memory requirements for resource-constrained applications. Resource-efficient ECG classification can be integrated into affordable wearable medical devices, making them accessible to populations across various income levels, and providing significant societal and economic impacts \cite{CVD21, UpBeat}. In this paper, we present a memory-efficient approach for ECG heartbeat classification. 

The main contributions of this work are as follows: 

\begin{itemize}
    \item A multi-feature fusion approach is proposed to combine the \textit{time interval} and \textit{under-the-curve area} features extracted from the ECG signal, to enhance accuracy and robustness of the model. This technique significantly improves accuracy, especially for the challenging RBBB and LBBB heartbeat classes.
    \item A Bi-LSTM based network is developed for heartbeat classification, achieving higher accuracy and 28\% less network parameters compared to the conventional LSTM network. 
    \item A memory-efficient classification approach is adopted which results in smaller networks with descent accuracy \textit{across all classes}. Multiple neural networks with varying parameter sizes, including tiny (T), small (S), medium (M), and large (L) models, are developed and compressed using various post-training quantization techniques to achieve state-of-the-art performance. The compressed L model with 8-bit integer quantization (INT8) achieves an accuracy of 88.4\% with 1.3\,MB memory. The compressed T model with dynamic range quantization (DRQ) can achieve 94.6\% accuracy with only 139\,kB memory.
\end{itemize}

The paper is structured as follows. In Section II, data preparation and signal preprocessing are discussed. In Section III, the multi-feature fusion technique is presented. Classification approach is discussed in Section IV. Details of the model training and test are presented in Section V, and model compression is discussed in Section VI. A comparison with state-of-the-art is presented in Section VII.

\section{Data Preparation}

\subsection{Dataset}

\begin{figure}[!t]
  \centering
  \includegraphics[width = 1.0\columnwidth]{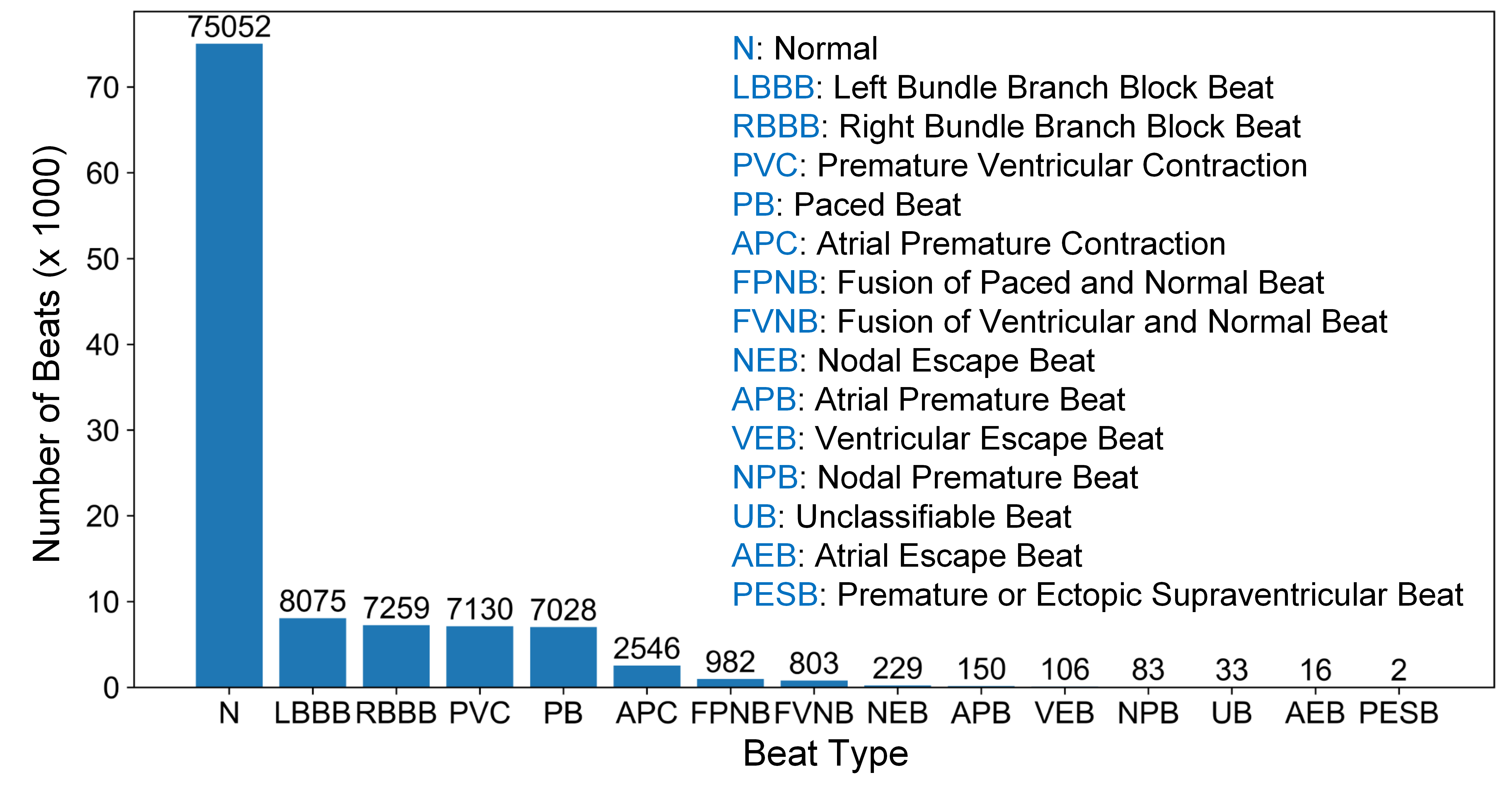}
  \caption{Distribution of data annotations in the MIT-BIH Arrhythmia Database.}
  \label{fig:database}
\end{figure}

The MIT-BIH Arrhythmia Database \cite{Moody01} consists of 48 ECG records, each 30 minutes in duration, sampled at a rate of 360 Hz.
%Each recording includes two leads: lead II obtained from the electrode on the limbs and lead V1 obtained from the electrode on the chest. Subjects were chosen from 4000 24-hour ambulatory ECG recordings at Boston’s Beth Israel Hospital, with 60\% inpatients and 40\% outpatients. 
In this work, the database records are partitioned as two-thirds for training and one-third for testing. We extracted the number of different arrhythmia types in the dataset, as shown in Fig. \ref{fig:database}. The normal heartbeat (N) and the four most common arrhythmia heartbeats, Right Bundle Branch Block (RBBB), Left Bundle Branch Block (LBBB), Premature Ventricular Contraction (PVC), and Paced Beat (PB), are selected for classification. In future work, including more classes can provide more clinically relevant results, but this will necessitate the use of a more complex neural network, which complicates resource-efficient implementation. 

RBBB, while benign in healthy subjects, has been associated with an increased risk of heart failure and atrial fibrillation in patients. LBBB, less common in healthy individuals than RBBB, typically indicates underlying cardiac issues and is linked with a higher risk of serious cardiac events such as heart failure and sudden cardiac death in CVD patients \cite{lilly2012braunwald}. Consequently, detecting these classes accurately is of \textit{vital importance}. However, many previous works have combined these two classes with the normal class. PVC is generally benign in healthy individuals but can signal an increased risk of cardiac death in CVD patients. PB is typically observed in patients with an active pacemaker. We use the MIT-BIH standard as it can provide more accurate medical information compared to the AAMI standard \cite{AAMI98}. In the AAMI standard, heartbeats are labeled into four main classes: Normal (N), Supraventricular Ectopic Beat (SVEB), Ventricular Ectopic Beat (VEB), and Fusion (F) \cite{AAMI98}. The normal class comprises the MIT-BIH classes N, RBBB, and LBBB. However, the significant risks associated with RBBB and LBBB indicate that this can overlook critical medical conditions.

\subsection{Preprocessing}

The recorded ECG signal is often distorted by noise and artifacts, which can hinder the extraction of key morphological features in the ECG waveform. Specifically, features with lower signal power, such as the P peak and Q dip, can be significantly affected by noise. Therefore, it is essential to mitigate this noise through preprocessing. The spectrum of noise and distortion signals is time-dependent, rendering conventional Fourier transform approaches ineffective for filtering. Instead, the discrete wavelet transform (DWT) can be used to suppress noise and distortion signals, as detailed in \cite{Elgendi10, aziz21}.

The Daubechies 4 (DB4) wavelet, which closely resembles the ECG signal waveform, is utilized in this work. 
This wavelet features a normalized central frequency $k_c \approx 0.7$. The pseudo-frequency for scale $n$ is calculated as $f_n = k_c f_s/2^n$, where $f_s$ is the sampling frequency of the ECG signal (360 Hz) \cite{aziz21}. The low-frequency noise, primarily baseline drift, is mostly concentrated around 0.5 Hz. Therefore, to achieve $f_n \leq 0.5$\,Hz, the signal decomposition should be performed up to the scale 9. Spectrum of the ECG signal spans a bandwidth of 0.05--100 Hz. However, most of the signal power lies within the 8--20 Hz band. It has been demonstrated that decomposition up to level 6 is necessary to capture this frequency band effectively \cite{Elgendi10}. High-frequency noise, often due to random body movements, usually is spread at frequencies above 50 Hz \cite{aziz21}. This noise is attenuated by discarding the detailed coefficients of levels 1, 2, and 3. Therefore, the signal is reconstructed using the detailed coefficients of levels 4, 5, and 6, along with the approximate coefficient of level 9. 
%The raw and filtered ECG signals are shown in Fig. \ref{fig:ECG_signals}.

%\begin{figure}[!t]
 %\centering
 % \includegraphics[width = \columnwidth]{figs/ECG_signals_0.png}
 % \caption{The raw and reconstructed filtered ECG signal waveforms.}
 % \label{fig:ECG_signals}
%\end{figure}

\section{Multi-Feature Fusion}

\subsection{Proposed Approach}
Multi-feature fusion involves combining various types of data or features from multiple sources or modalities to enhance classification accuracy. Features can encompass diverse forms such as images, text, audio, video, or different metrics extracted from the same dataset. In this paper, we extract two sets of features from the dataset, time intervals $T_{ij}$ and under-the-curve areas $A_{ij}$, from the main points (PQRST) of ECG signal waveform. These two sets, comprising six time intervals and four under-the-curve areas, are integrated using an early fusion approach, and are fed into the classification pipeline.

The extracted time intervals include $T_{\rm RR}$, $T_{\rm PR}$, $T_{\rm RT}$, $T_{\rm QR}$, $T_{\rm RS}$, and $T_{\rm PT}$. The RR interval is defined as the time difference between the T peaks of consecutive heartbeats. The other time intervals, $T_{ij}$, are defined as the time differences between two main points within a single heartbeat 
\begin{equation}
    \label{time_features}
    T_{ij} = t_j - t_i.
\end{equation}

The extracted under-the-curve areas include $A_{\rm PQ}$, $A_{\rm ST}$, $A_{\rm QR}$, and $A_{\rm RS}$. These features are calculated as the sum of the absolute values of the signal samples located between two main points, inclusive of the points themselves, expressed as 
\begin{equation}
    \label{area_features}
    A_{ij} = \sum_{k = i}^{j} {|x(t_k)|}.
\end{equation}
An important advantage of the under-the-curve area features is their \textit{robustness against noise}, as a result of the inherent averaging property of the summation function.

For each of the main points in the ECG waveform, a specific algorithm should be employed to achieve high accuracy. The MIT-BIH dataset provides standard annotations for only the R and P peaks, which facilitates the evaluation of their detection accuracy. However, for the other main points, accuracy estimation is not possible. The effectiveness of these detection algorithms can ultimately be validated through the overall accuracy of the heartbeat classification.

\subsection{Detection of R Peaks}

\begin{figure}[!t]
  \centering
  \includegraphics[width = \columnwidth]{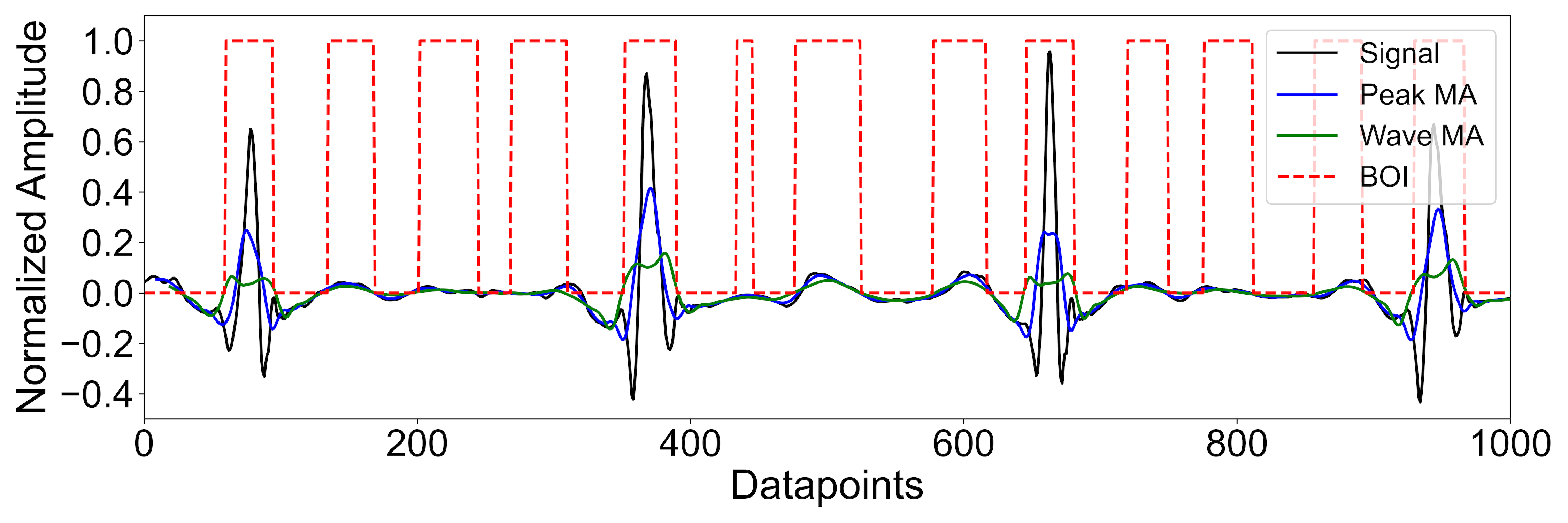}
  \caption{ECG signal, peak and wave moving averages, and block of interest (BOI) for detection of R peaks.}
  \label{fig:moving_averages}
\end{figure}

We use two moving average windows to implement adaptive blocks for the accurate detection of the R peaks:
\begin{equation}
    \label{MA1}
    y_p[n] = \frac{1}{2N+1} \sum_{k=-N}^{N} x[n+k]
\end{equation}
\begin{equation}
    \label{MA2}
    y_w[n] = \frac{1}{2M+1} \sum_{k=-M}^{M} x[n+k].
\end{equation}
The first moving average (peak) $y_p[n]$ is used to identify the QRS complex and the second moving average (window) $y_w[n]$ serves as a threshold for the first. It is important that the length of the first window, $N$, is shorter than that of the second window, $M$, to maintain effective detection. Specifically, the first window's length is set at 36 samples (100 ms), corresponding to the duration of a typical QRS complex. A threshold of 0.3 is applied to this window to exclude noise that could mistakenly be identified as the QRS complex. The length of the second window is determined by the typical duration of a normal heartbeat, set at 120 samples (330 ms). A Block of Interest (BOI) is defined such that it equals 1 when $y_p[n] > y_w[n]$ and 0 otherwise. This function generates a sequence of rectangular pulses with adaptive widths, which are then used to detect the R peaks. In Fig. \ref{fig:moving_averages}, samples of the ECG signal, peak and wave moving averages, and BOI are shown. This approach resulted in the sensitivity and precision higher than 90\% across the records. 

\subsection{Detection of P and T Peaks}

To detect the P and T peaks, a similar approach using dual moving average windows is employed, as with the R peak detection. The significantly higher amplitude of R peaks compared to the P and T peaks presents a detection challenge. To address this, the QRS complex is removed from the signal by zeroing a specific number of samples both before and after each R peak. Specifically, 30 samples preceding and 60 samples following the R peak are set to zero, based on the typical duration of the QRS complex in a normal heartbeat. 

The dual moving averages method is applied to the signal with the QRS complex removed, which allows for the detection of both P and T peaks. However, these peaks must be distinguished using additional processing. The length of the peak moving average is determined based on the duration of the P wave in a normal heartbeat, which is 20 samples (55 ms). Similarly, the length of the wave moving average is set according to the normal QT interval at 40 samples (110 ms). To distinguish between the P and T peaks, the method calculates the distance from the maximum point of each detected peak to the nearest R peak \cite{Elgendi10}. For P peaks, the distance to the closest R peak, $T_{\rm PR}$, ranges from 20 to 170 samples (55 to 470 ms). For T peaks, the distance to the R peak is between 40 and 210 samples (110 to 583 ms). Peaks are classified as either P or T based on these criteria. 
%This method, allowing an error tolerance of 36 samples (100 ms), achieved a sensitivity of 84.6\% for detecting P peaks. However, the accuracy for T peaks could not be assessed due to the absence of reference annotations.

\subsection{Detection of Q and S Dips}

The Q and S dips in the ECG signal waveform are detected using a similar dual moving averages approach. Unlike the P and T peaks, the QRS complex is retained in the signal. The detected R peaks serve as reference points for identifying the Q and S dips. These two dips are located on either side of the R peak at varying distances, which facilitates their detection. For the Q point, the maximum distance from the R peak, $T_{\rm QR}$, is 20 samples (55 ms). Similarly, for the S point, the maximum distance from the R peak, $T_{\rm RS}$, is 40 samples (110 ms).

\subsection{Accuracy of Detections}
\begin{figure}[!t]
  \centering
  \includegraphics[width =  \columnwidth]{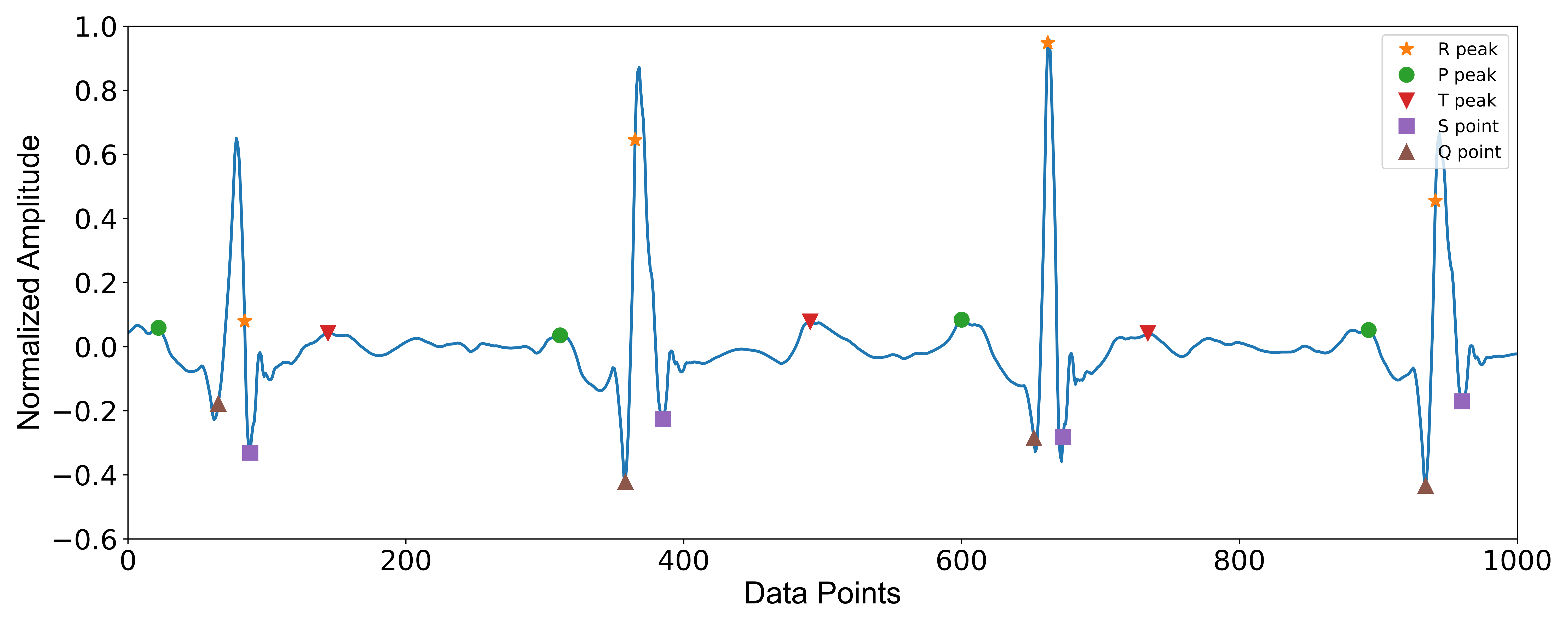}
  \caption{A sample of detected main points (PQRST) in the ECG signal.}
  \label{fig:detected_points}
\end{figure}

 A sample of the detected main points in the ECG signal is shown in Fig. \ref{fig:detected_points}.
 The sensitivity and precision for the R peaks are very high at 99.8\% and 99.95\%, respectively. For the P peaks, these metrics are recorded at 84.0\% and 84.6\%, respectively. However, the detection accuracy of other main points cannot be evaluated due to the absence of reference annotations in the MIT-BIH database.

\section{Heartbeat Classification}

\subsection{Classification Approach}

We pursue a design approach to achieve high accuracy \textit{across all classes}, rather than an \textit{overall accuracy}. The class N usually achieves the highest accuracy, while the RBBB and LBBB classes present the greatest challenges for precise classification. The overall accuracy is typically dominated by the class N and, as a result, the inaccuracies in other classes can be masked.
The medical implications of this approach are more useful than using only the overall accuracy which is popular in machine learning research \cite{lilly2012braunwald}. 

\begin{figure*}[!t]
  \centering
  \includegraphics[width = 1.8 \columnwidth]{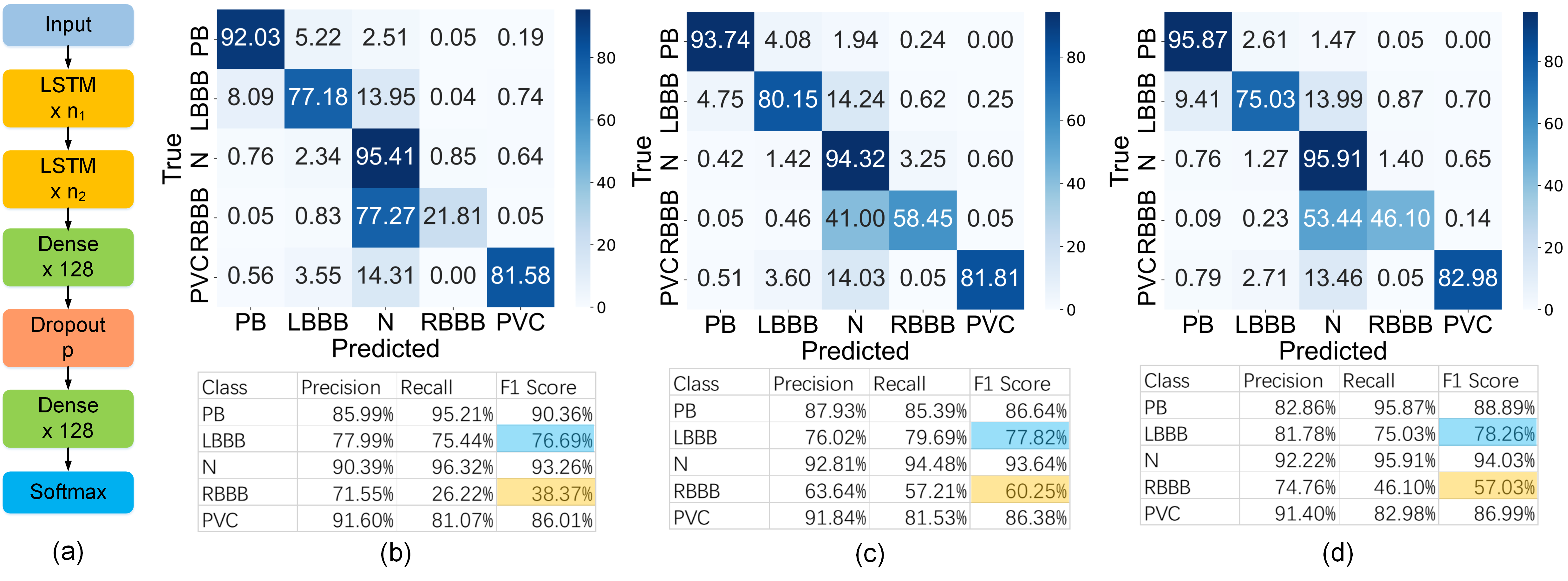}
  \caption{(a) Neural network architecture using LSTM layers and the six time interval features, (b) classification results for $n_1 = n_2 = 64$ and $p=0.5$, (c) classification results for $n_1 = n_2 = 128$ and $p=0.5$, (d) classification results for $n_1 = n_2 = 64$ and $p=0.25$.}
  \label{fig:LSTM_network}
\end{figure*}

\begin{figure*}[!t]
  \centering
  \includegraphics[width = 1.8 \columnwidth]{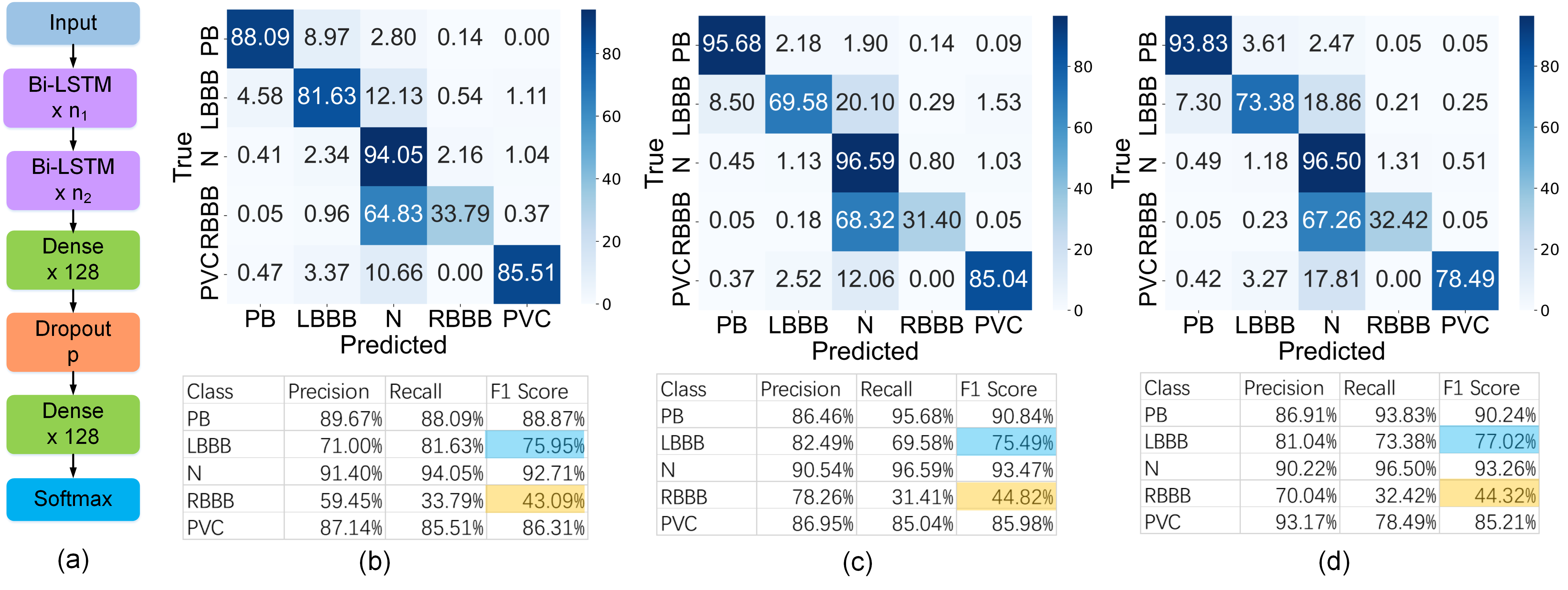}
  \caption{(a) Neural network architecture using Bi-LSTM layers and the six time interval features, (b) classification results for $n_1 = n_2 = 32$ and $p=0.5$, (c) classification results for $n_1 = n_2 = 64$ and $p=0.5$, (d) classification results for $n_1 = n_2 = 32$ and $p=0.25$.}
  \label{fig:BiLSTM_network}
\end{figure*}

\subsection{LSTM Networks}

In the preliminary simulations discussed in this subsection, we exclusively used the six time interval features ($T_{ij}$) to provide deeper insights. All models were implemented, trained, and tested using Python. We employed the sparse categorical cross-entropy loss function in the algorithms. The network architecture, depicted in Fig. \ref{fig:LSTM_network}(a), includes two LSTM layers, two dense layers, and a dropout layer \cite{Srivastava14}. We investigated various networks built using this configuration, adjusting the number of units, $n_1$ and $n_2$, in the two LSTM layers, as well as the dropout rate, $p$. The classification results for three scenarios are presented in Figs. \ref{fig:LSTM_network}(b)-(d).

Fig. \ref{fig:LSTM_network}(b) shows a low accuracy of 21.81\% for the RBBB class in the confusion matrix. Here, the precision metric is significantly higher than the recall metric (71.55\% vs 26.22\%), resulting in an F1 score of 38.37\%. 

In the first effort to improve accuracy, the number of LSTM units are doubled. The results shown in Fig. \ref{fig:LSTM_network}(c) indicate that the accuracy for the RBBB class is increased (accuracy of 58.48\% and F1 score of 60.25\%), though the accuracy and F1 scores for the LBBB and PVC classes showed only minor improvements. This suggests that a substantial increase in the number of LSTM units is necessary to significantly improve accuracy for these classes. However, such an approach results in a larger model, which conflicts with our goal of resource-efficient implementation. 
Fig. \ref{fig:LSTM_network}(d) shows that by changing the dropout rate from 0.5 to 0.25, the model achieves higher accuracy across all classes, especially for the RBBB class.

\subsection{Bi-LSTM Networks}

Intuitively, in a sequence of human heartbeats, each signal is correlated with its preceding and succeeding heartbeats. Thus, a Bi-LSTM network, which computes neuron outputs using both past (backward) and future (forward) states \cite{Schuster97}, has the potential to achieve higher accuracy. However, training Bi-LSTM networks can be challenging due to the presence of dual signal flow paths. Fig. \ref{fig:BiLSTM_network}(a) shows the classification network, which comprises Bi-LSTM, dense, and dropout layers. This network has the same architecture as the LSTM based network shown in Fig. \ref{fig:LSTM_network}(a), with the exception that LSTM layers are replaced by Bi-LSTM layers to allow a fair comparison. Additionally, only the six time interval features are used as inputs to the network. 

The classification results for three representative networks are depicted in Fig. \ref{fig:BiLSTM_network}. Specifically, Fig. \ref{fig:BiLSTM_network}(b) shows the results for a network including Bi-LSTM layers with $n_1 = n_2 = 32$ units each, and a dropout layer with a 0.5 dropout rate. The number of units in the Bi-LSTM network is set at half that of the LSTM network to ensure both networks have approximately the same number of parameters. Consequently, this Bi-LSTM-based network comprises about 42.5k parameters, compared to a similar LSTM-based network with 64 units which comprises about 58.9k parameters. Therefore, the Bi-LSTM based network achieves higher accuracy (33.79\% vs 21.81\%) and F1 score (43.09\% vs 38.37\%) for the RBBB class, while reducing the number of parameters by 28\%.

\subsection{Impact of Multi-Feature Fusion}

\begin{figure*}[!t]
  \centering
  \includegraphics[width = 1.7 \columnwidth]{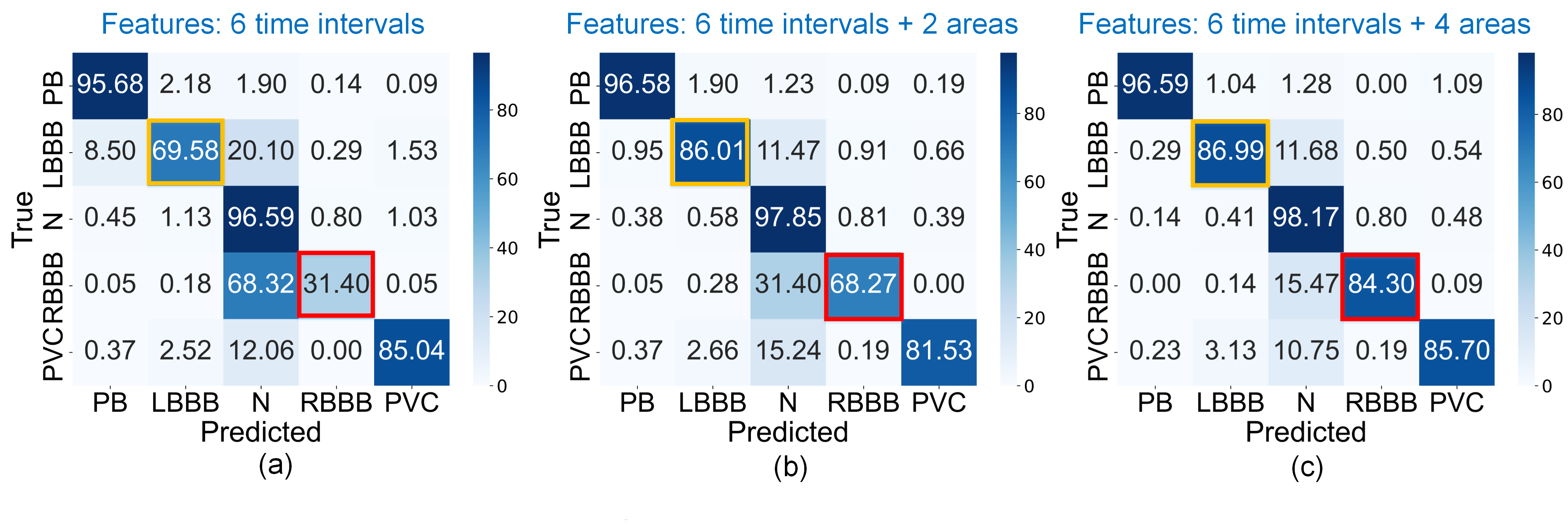}
  \caption{Confusion matrices for the Bi-LSTM based network with different input features. (a) six time interval features, (b) six time intervals + two area features, (c) six time intervals + 4 area features. The major improvements are achieved for the RBBB and LBBB classes.}
  \label{fig:feature_fusion}
\end{figure*}

The confusion matrices for three scenarios of the input features are shown in Fig. \ref{fig:feature_fusion}. The network architecture remains consistent with that depicted in Fig. \ref{fig:BiLSTM_network}(a), featuring two 64-unit Bi-LSTM layers and a dropout rate of 0.5. It is concluded that by incorporating two area features along with the six time intervals, \textit{the accuracy for the RBBB class significantly increases} from 31.40\% to 68.27\%. Similarly, the accuracy for the LBBB class improves from 69.58\% to 86.01\%. Using the multi-feature fusion comprising six time intervals and four area features, the accuracy for the RBBB class reaches 84.30\%. Furthermore, the accuracy for all other classes also shows improvement over using only the time interval features.

\begin{figure*}[!t]
 \centering
 \includegraphics[width = 1.7\columnwidth]{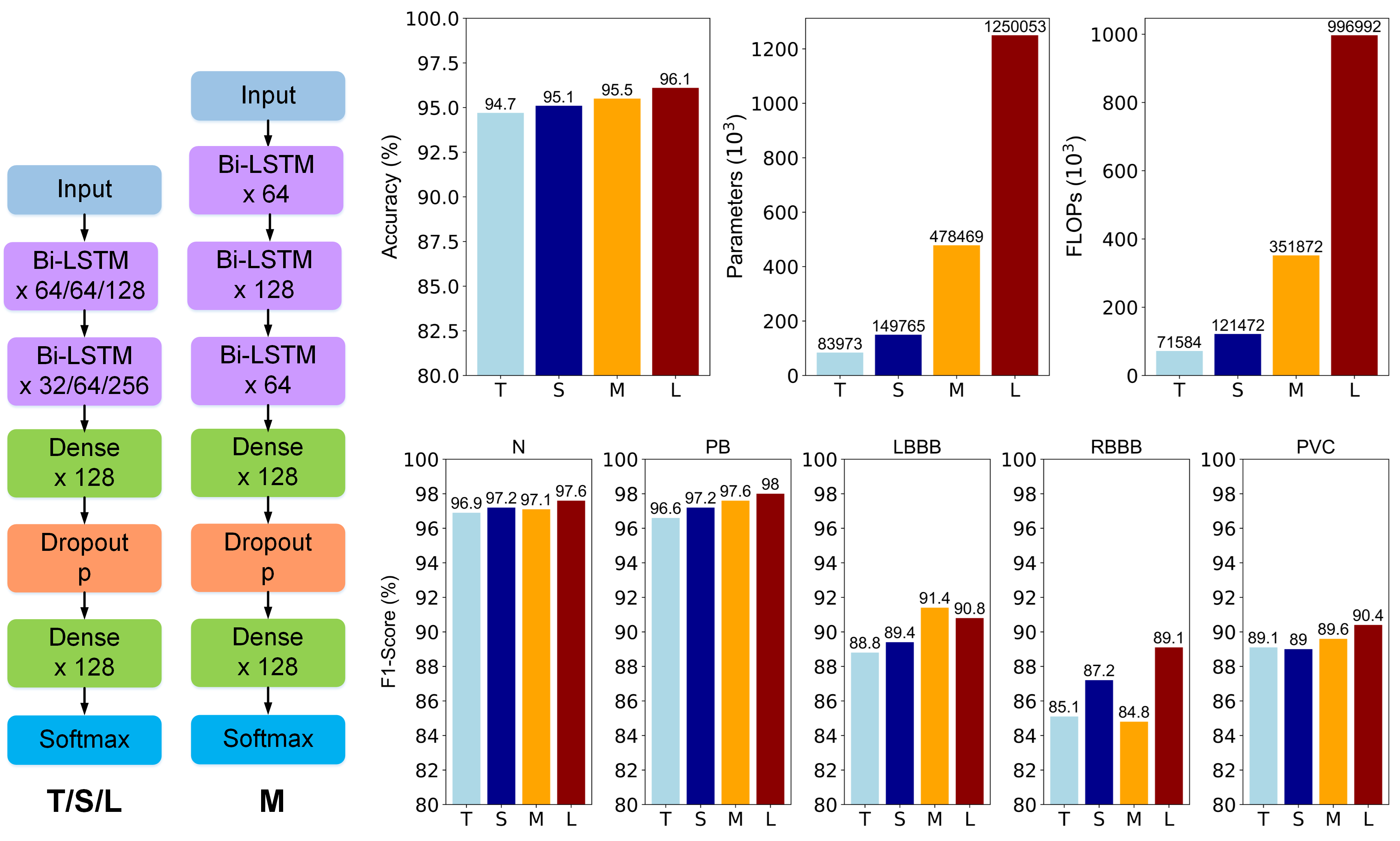}
  \caption{Neural network architecture, accuracy, the number of parameters, the number of FLOPs, and F1 score across heartbeat classes for the four developed models, tiny (T), small (S), medium (M), and large (L).}
  \label{fig:4models}
\end{figure*}

\section{Model Training and Test}

\subsection{Training Results}

All networks apply a multi-feature fusion strategy, incorporating six time intervals and four under-the-curve areas. In all of the LSTM and Bi-LSTM layers, the tanh activation function is used. Activation function used in the first dense layer is ReLU and in the second dense layer, the softmax function is used. The four developed models are trained using mini-batch gradient descent with a batch size of 64. The number of training samples is 73180. Each epoch takes 1143 iterations. The sparse categorical cross-entropy loss function is used in training the models 
\begin{equation}
\label{loss_fucntion}
L = - \frac{1}{N}\sum_i \log (p_i).
\end{equation}
The loss is computed using the TensorFlow package in Python. 
%In Fig. \ref{fig:training_test_loss}, training and test losses are shown for the four models.
All models are trained up to an epoch number of 10, which roughly provides a good compromise between training and test loss without overfitting. 

%\begin{figure}[!t]
%  \centering
% \includegraphics[width = \columnwidth]{figs/training_test_loss.png}
%  \caption{Training and test losses of the developed four models.}
%  \label{fig:training_test_loss}
%\end{figure}

\subsection{Test Results}

Leveraging the insights from preliminary investigations, we developed and evaluated several models to achieve high accuracy across all classes while maintaining a small model size. These models are realized using multiple Bi-LSTM layers, with the number of units in each layer scaled to achieve optimal accuracy. The architecture of the four developed models is shown in Fig. \ref{fig:4models}(a).

Four models of varying sizes are presented as representatives of the developed models: a tiny model (T) with 84k parameters, a small model (S) with 150k parameters, a medium model (M) with 478k parameters, and a large model (L) with 1.25M parameters. The performance of these models is summarized in Table \ref{tab:4models}. Fig. \ref{fig:4models} compares the neural network architecture, overall accuracy, number of parameters, memory, number of FLOPs, and F1 score for each heartbeat class across the four models. Notably, there are only slight variations in overall accuracy between models of significantly different sizes. F1 score for class N is nearly consistent across all models and closely follows the overall accuracy as shown in Fig. \ref{fig:4models}. For the challenging RBBB class, the best F1 score is achieved with the large model, while the tiny model exhibits the lowest F1 score.

\begin{figure}[!t]
  \centering
    \includegraphics[width = \columnwidth]{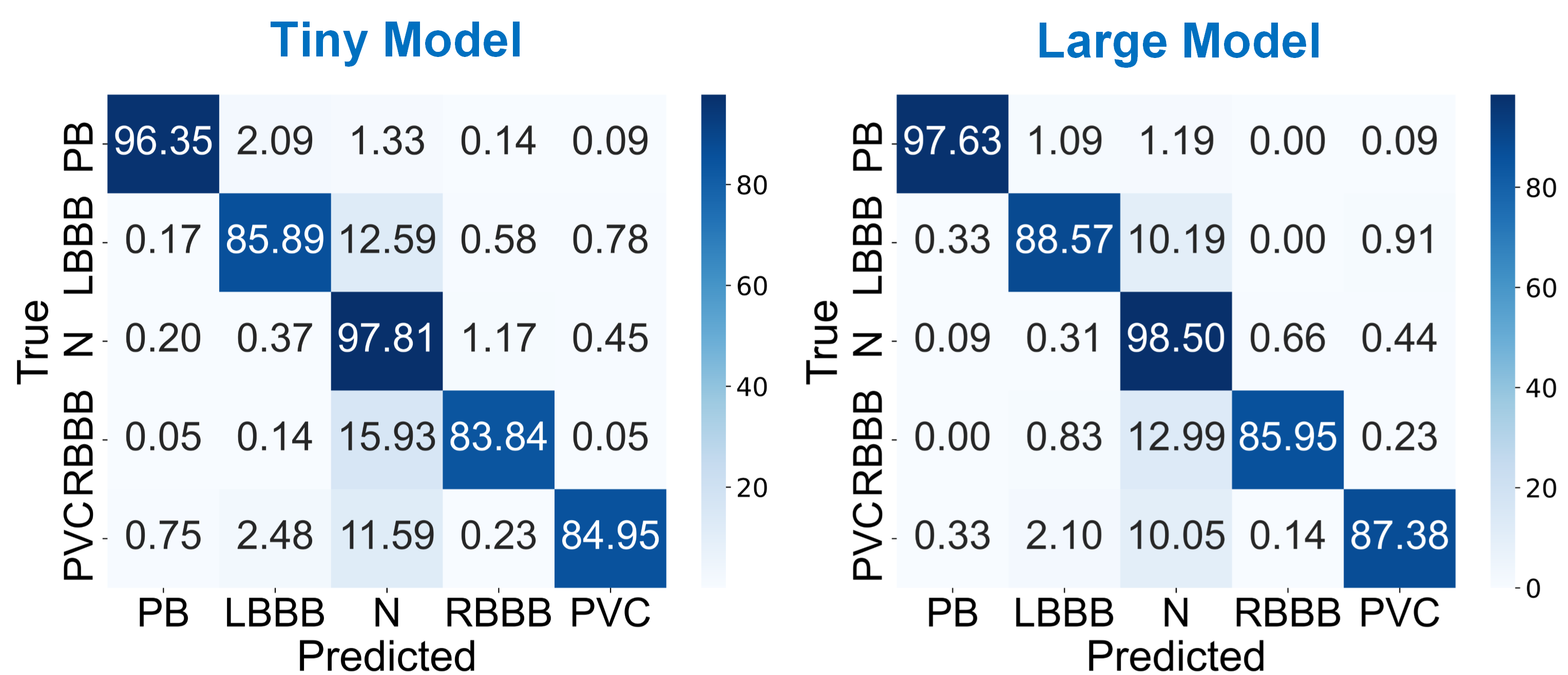}
  \caption{Confusion matrix for the developed tiny (T) and large (L) models.}
  \label{fig:2models_confusion}
\end{figure}

The confusion matrices for the tiny (T) and large (L) models are shown in Fig. \ref{fig:2models_confusion}. The matrices are predominantly diagonal with small non-diagonal elements, indicating that both models are effective in distinguishing between the classes. 

\begin{figure}[!t]
  \centering
  \includegraphics[width = 1.0 \columnwidth]{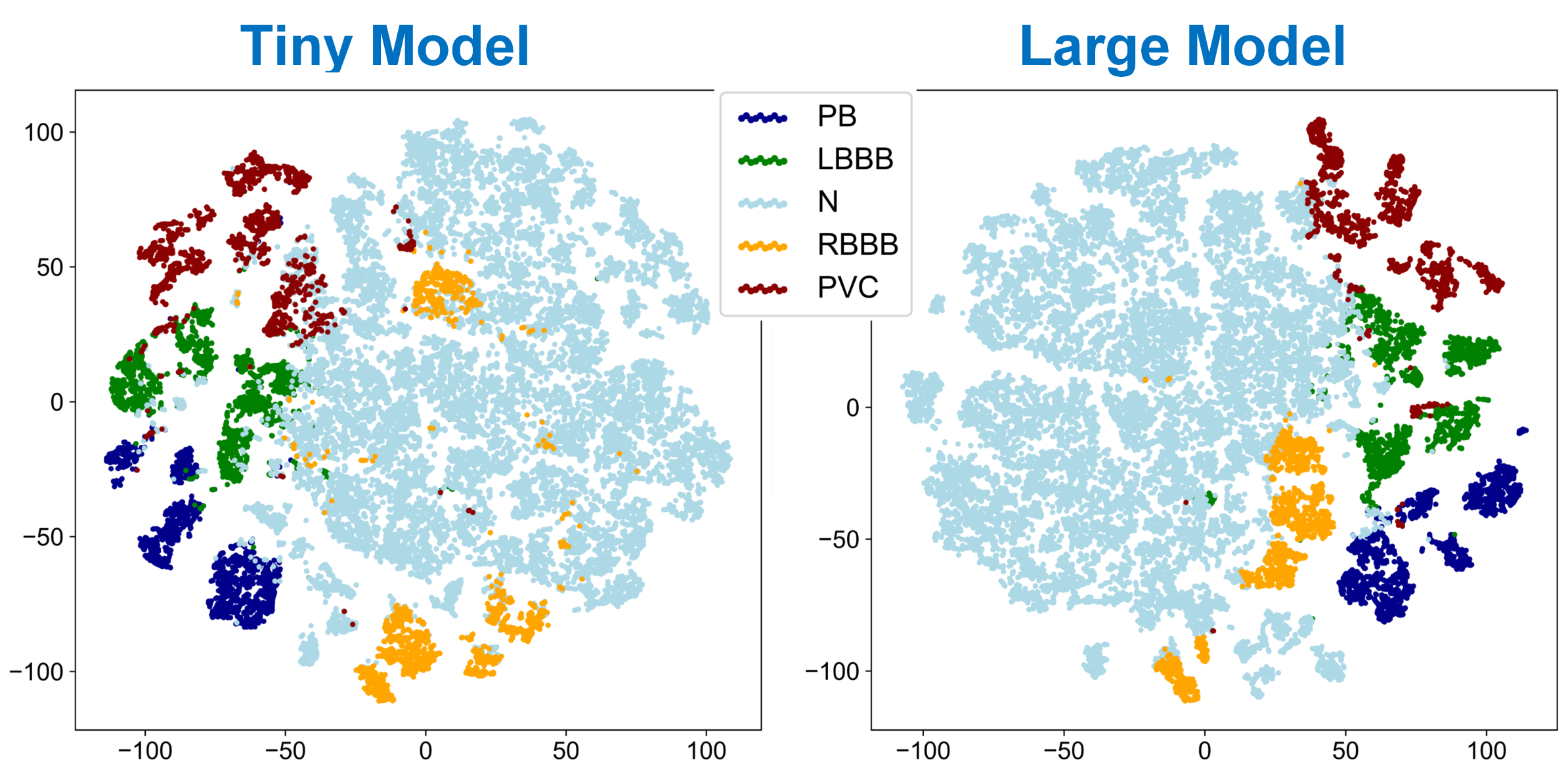}
  \caption{t-SNE visualization for the developed tiny (T) and large (L) models. The L model is more effective in clustering samples of various classes.}
  \label{fig:2models_TSNE}
\end{figure}

The t-SNE visualizations \cite{TSNE08} for the two models are shown in Fig. \ref{fig:2models_TSNE}. The models have effectively clustered most of the data samples, with only minor exceptions in the tiny model and a few points in the large model. The challenges of developing these models for dataset classification are evident in the t-SNE plots, such as, the large size of the N class and the close proximity of the difficult RBBB class to the N class. \textit{The t-SNE plots also confirm the high capabilities of the large model} to effectively separate the heartbeat classes, which \textit{was not clear from the accuracy and F1 score metrics}.

\begin{table*}[!t]
  \caption{Performance summary of the four developed models (before compression).}
  \label{tab:4models}
  \centering
  \renewcommand{\arraystretch}{1.25}
   \setlength{\tabcolsep}{8pt}
  \begin{tabular}{cccccccccc}
    \hline
& & \multicolumn{5}{|c|}{\textbf{F1 Score (\%)}} & \multicolumn{3}{c}{\textbf{Model Size}}\\ 
        \hline
        \hline
        \textbf{Model}  & \textbf{Acc (\%)} & \textbf{N} & \textbf{PB} & \textbf{LBBB} & \textbf{RBBB} & \textbf{PVC} & \textbf{Param} & \textbf{Memory} &  \textbf{FLOPs}\\ 
        \hline 
        Tiny & 94.7 & 96.9 & 96.6 & 88.8 & 85.1 & 89.1 & 83973 & 328 kB & 71584 \\
        \hline 
        Small  & 95.1 & 97.2 & 97.2 & 89.4 & 87.2 & 89 & 149765 & 585 kB & 121472 \\
        \hline 
        Medium  & 95.5 & 97.1 & 97.6 & 91.4 & 84.8 & 89.6 & 478469 & 1.83 MB & 351872 \\
        \hline 
        Large  & 96.1 & 97.6 & 98 & 90.8 & 89.1 & 90.4 & 1250053 & 4.77 MB & 996992 \\
    \hline
  \end{tabular}
\end{table*}

\section{Model Compression}

\begin{figure*}[!t]
  \centering
\includegraphics[width = 1.6 \columnwidth]{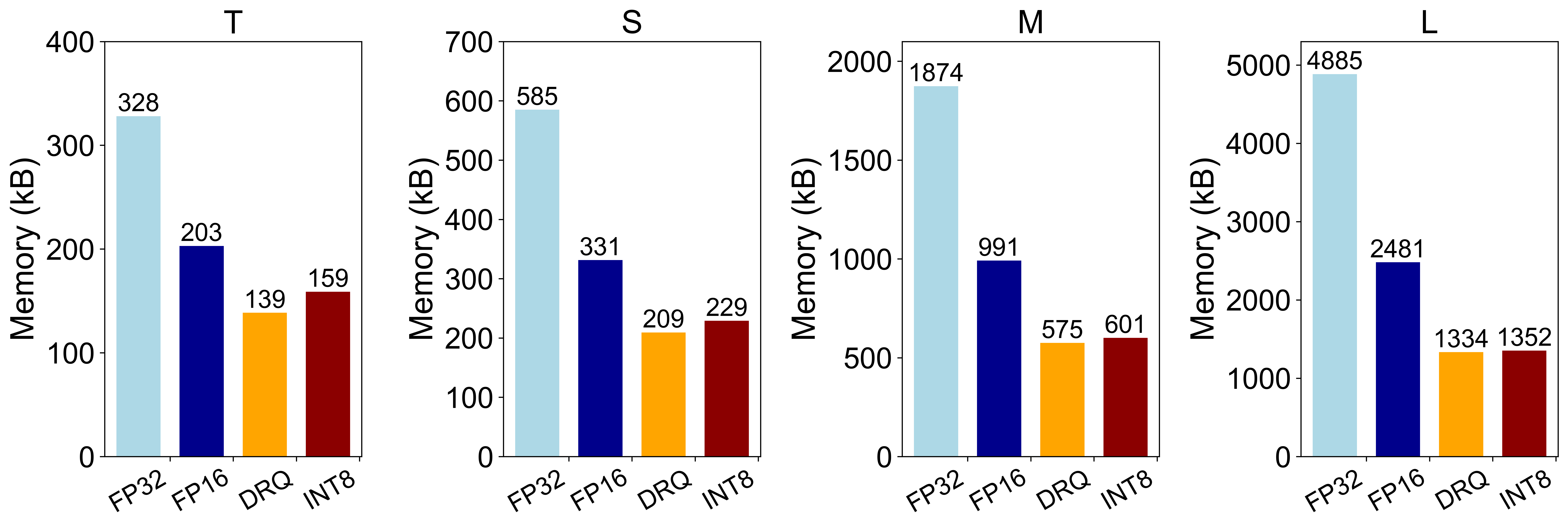}
  \caption{Memory of the developed models for various quantization methods. DRQ: dynamic range quantization.}
  \label{fig:memory_quantization}
\end{figure*}

\begin{figure*}[!t]
  \centering
\includegraphics[width = 1.6 \columnwidth]{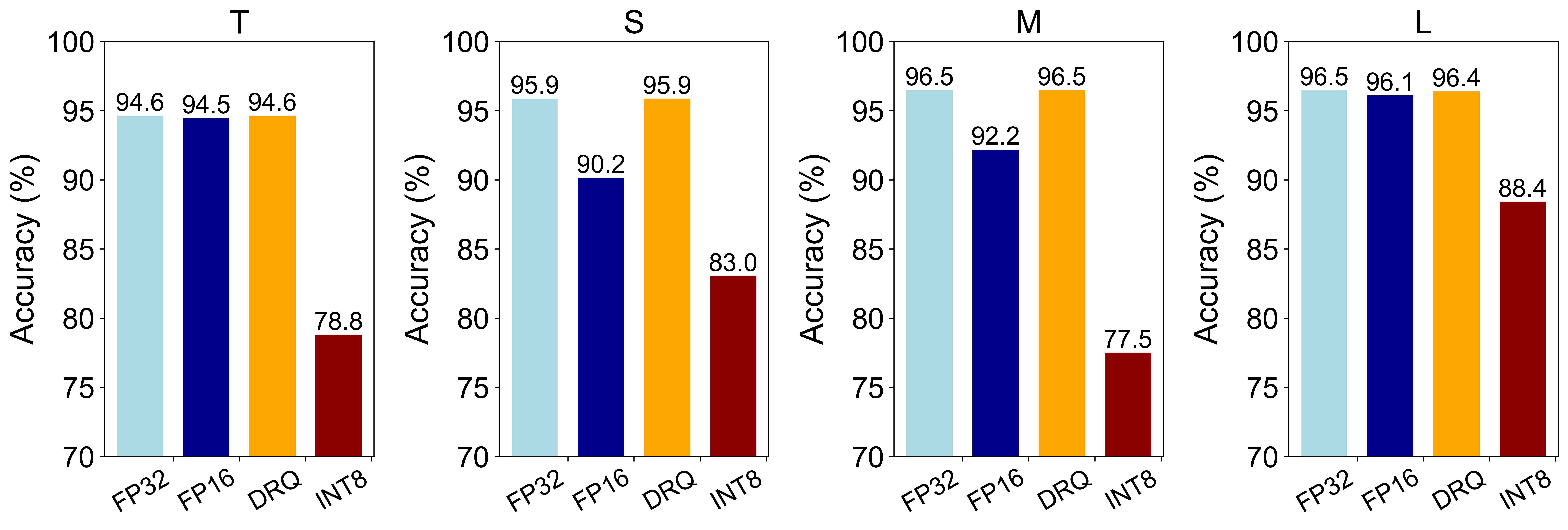}
  \caption{Accuracy of the developed models for various quantization methods. DRQ: dynamic range quantization.}
  \label{fig:accuracy_quantization}
\end{figure*}

The developed neural networks are compressed through various post-training quantization techniques to evaluate their accuracy-memory trade-offs for resource-constrained devices \cite{gholami-2021, han-2015}. We use the 16-bit floating-point (FP16) format, the 8-bit full integer quantization (INT8), and the dynamic range quantization (DRQ). Memory and accuracy of the developed four models for various quantization methods are shown in Fig. \ref{fig:memory_quantization} and Fig. \ref{fig:accuracy_quantization}. 

In the FP16 format, the weights and activations are converted to an half precision of the original FP32 format. This method can reduce the memory size by up to 50\% without significant loss of the accuracy. Fig. \ref{fig:memory_quantization} shows that this approach reduces the memory to 203\,kB for the T model (0.62x), 331\,kB for the S model (0.57x), 991\,kB for the M model (0.53x), and 2481\,kB for the L model (0.51x). In Fig. \ref{fig:accuracy_quantization}, the difference between accuracy of the FP16 and FP32 formats is 0.1, 5.7, 4.3, and 0.4 percentage point, for the T, S, M, and L models, respectively. Therefore, the model compression ratio is limited to 2 and the accuracy drop is small in this approach.

In the full integer quantization, the weights and activations are mapped from the FP32 format to an integer in the range [$-$127, 128] for 8-bit integers (INT8). 
%This approach can further compress the model. However, it leads to significant accuracy losses and increased quantization complexity. 
Fig. \ref{fig:memory_quantization} shows that using the INT8 quantization, the memory is reduced to 159\,kB for the T model (0.48x), 229\,kB for the S model (0.39x), 601\,kB for the M model (0.32x), and 1352\,kB for the L model (0.28x). The compression ratio is on the order of 2--4, and it is higher for the L model. However, the accuracy shown in Fig. \ref{fig:accuracy_quantization} is dropped by 10--20 percentage point compared to the original FP32 results. The highest accuracy of 88.4\% is achieved for the compressed L model, which also benefits from the largest compression ratio. The memory required for the compressed L model (about 1.3\,MB) allows the model to be realized on a commercial microcontroller unit (MCU) and be integrated in a wearable device. 

In the dynamic range quantization (DRQ), the weights are statically quantized from FP32 to INT8 while the activations are stored in FP32 during the conversion time. The activations are dynamically quantized to INT8 during runtime. Therefore, this approach can potentially achieve both small memory requirement and high accuracy. In Fig. \ref{fig:memory_quantization} it is noted that the memory requirement of the DRQ is very close to that of the full integer quantization (INT8). Interestingly, accuracy of the DRQ approach as shown in Fig. \ref{fig:accuracy_quantization} is the same as the original FP32 model. The compressed T model can achieve an accuracy of 94.6\% with only 139\,kB memory. For the compressed L model, the accuracy is 96.4\% and the required memory is 1.33\,MB.

In Table \ref{tab:quantization}, F1 score is presented across heartbeat classes, developed models, and quantization methods. General trend is consistent with the presented discussions: the dynamic range quantization achieves an accuracy very close to the original FP32, the accuracy of FP16 is lower than FP32 by a few percentage point, and INT8 features the lowest accuracy. RBBB and LBBB heartbeat classes, which were difficult to precisely classify in the original models, are also sensitive to the quantization. This can be addressed by retraining the neural network and testing the model accuracy after quantization. 

\begin{table}[!t]
  \caption{F1 Score across heartbeat classes, developed models, and quantization methods.}
  \label{tab:quantization}
  \centering
  \renewcommand{\arraystretch}{1.25}
   \setlength{\tabcolsep}{6pt}
  \begin{tabular}{cccccccccc}
    \hline
 & & \multicolumn{5}{c}{\textbf{F1 Score (\%)}} \\
\cmidrule(lr){3-7}
\textbf{Model} & \textbf{Quant.} & \textbf{N} & \textbf{PB} & \textbf{LBBB} & \textbf{RBBB} & \textbf{PVC} \\
    \hline \hline
\multirow{5}{*}{Tiny} & FP32   & 96.8 & 96.9 & 88.9 & 80.4 & 89.4\\
                      & FP16 & 89.3 & 83.8 & 66.1 & 27.3 & 76.1\\
                      & DRQ    & 96.8 & 96.9 & 88.9 & 80.5 & 89.4\\
                      & INT8 & 86.9 & 73.2 & 50.2 & 12.7 & 74.9\\
    \hline
\multirow{5}{*}{Small} & FP32   & 97.4 & 97.6 & 91.9 & 88.5 & 90.4\\
                      & FP16 & 95.1 & 94.4 & 84.6 & 75.3 & 85.5\\
                      & DRQ    & 97.4 & 97.5 & 91.9 & 88.5 & 90.4\\
                      & INT8 & 89.5 & 87.6 & 62.2 & 27.0 & 74.6\\
    \hline
\multirow{5}{*}{Medium} & FP32   & 97.8 & 98.0 & 93.1 & 90.3 & 90.8\\
                      & FP16 & 97.5 & 97.4 & 91.6 & 87.8 & 90.8\\
                      & DRQ    & 97.8 & 98.0 & 93.1 & 90.3 & 90.9\\
                      & INT8 & 86.4 & 61.6 & 48.6 & 36.4 & 69.5\\
    \hline
\multirow{5}{*}{Large} & FP32   & 97.7 & 98.0 & 92.8 & 90.0 & 91.3\\
                      & FP16 & 97.6 & 97.6 & 92.2 & 88.3 & 91.1\\
                      & DRQ    & 97.8 & 98.0 & 92.6 & 89.6 & 91.3\\
                      & INT8 & 92.6 & 91.3 & 77.2 & 51.5 & 83.4\\
     \hline
    \end{tabular}
 \end{table}

\begin{table*}[!t]
 \caption{Comparison of the developed tiny (T) and large (L) models with state-of-the-art ($^*$ estimated).}
 \label{tab:comparison}
 \centering
 \renewcommand{\arraystretch}{1.25}
 \setlength{\tabcolsep}{7pt}
 \begin{tabular}{cccccccccccc}
  \hline
& & & & \multicolumn{5}{|c|}{\textbf{F1 Score (\%)}} & & \multicolumn{2}{|c}{\textbf{Memory (kB)}} \\
        \hline
        \hline
        \textbf{Ref} & \textbf{Features} & \textbf{Classifier} & \textbf{Acc (\%)} & \textbf{N} & \textbf{PB} & \textbf{LBBB} & \textbf{RBBB} & \textbf{PVC} & \textbf{Param} & \textbf{32-bit} & \textbf{8-bit}\\
        \hline
        \cite{kachuee18}  & ECG beats & Res CNN & 93.4 & 90.9 & --- & --- & --- & 92.5 & 99k & 396$^*$ & ---\\
       \hline
        \cite{alfaras19} & Raw ECG, $T_{\rm RR}$ & ESN & 97.9 & \multicolumn{5}{c}{overall F1 score: 89.7\%} & 30k & 120$^*$ & ---\\
        \hline
        \cite{aziz21} & $T_{\rm PR}$, $T_{\rm RT}$, age, sex & SVM & 82.2 & 68.4 & 100 & 13.0 & 100 & 100 & --- & --- & ---\\
        \hline
        \cite{aziz21} & $T_{\rm PR}$, $T_{\rm RT}$, age, sex & MLP & 80.0 & 77.9 & 85.1 & 76.5 & 74.8 & 91.6 & --- & --- & ---\\
        \hline
        %\cite{khan21} & PCA & LSTM & 93.5 & \multicolumn{5}{c}{overall F1 score: 91.7\%} & --- & ---& ---\\
        %\hline
        \cite{rajkumar19} & Raw ECG & CNN & 93.6 &--- & --- & ---& --- & --- &153k & 612$^*$ & ---\\
        \hline
        \cite{petmezas21} & Raw ECG & CNN-LSTM & 97.8 & 98.4 & --- & --- & --- & --- & --- & --- & ---\\
        \hline
        \cite{obeidat21} & Raw ECG & CNN & 97.4 & --- & --- & 95.0 & 96.9 & --- & 727k & 2908$^*$ & ---\\
        \hline
        \cite{obeidat21} & Raw ECG & LSTM & 97.1 & --- & --- & 94.8 & 96.4 & --- & 52k & 208$^*$& ---\\
        \hline
        \cite{obeidat21} & Raw ECG & CNN-LSTM & 98.2 & --- & --- & 97.0 & 97.9 & --- & 64k & 256$^*$ & ---\\
        \hline
        Model T & 6 $T_{ij}$, 4 $A_{ij}$ & Bi-LSTM & 94.7 & 96.9 & 96.6 & 88.8 & 85.1 & 89.1 & 84k & 328 & 139\\
        \hline
        Model L & 6 $T_{ij}$, 4 $A_{ij}$ & Bi-LSTM & 96.1 & 97.6 & 98.0 & 90.8 & 89.1 & 90.4 & 1.25M & 4885 & 1334 \\
        %\hline
        %This work & 6 $T_{ij}$, 4 $A_{ij}$ & Bi-LSTM 11 & 96 & 0.97 & 0.98 & 0.92 & 0.9 & 0.91 & 1496325 & & \\
     \hline
    \end{tabular}
 \end{table*}

\section{Comparison with State-of-the-Art}

In Table \ref{tab:comparison}, performance of the tiny (T) and large (L) models, as representatives of the developed models, are compared with state-of-the-art models. F1 score across the heartbeat classes is presented for a detailed comparison of the models. 

However, in the literature heartbeat classes are often defined based on varying criteria, making fair comparisons with these works challenging. For instance, \cite{kachuee18} defines the class N as encompassing the N, RBBB, LBBB, AEB, and NEB annotations from the MIT-BIH dataset (Fig. \ref{fig:database}). However, as we discussed earlier, it is crucial to separate these classes. Additionally, as can be noted in Table \ref{tab:comparison}, accuracy for individual classes has not been reported in the most of previous work.

The size of the neural networks is benchmarked using their number of parameters. The T model comprises 84\,k parameters with 94.7\% overall accuracy and F1 score higher than 85.1\% across all classes. The ESN (echo state network) presented in \cite{alfaras19} comprises 30\,k parameters which is smaller than other networks. However, this work has only presented the overall accuracy without details about individual classes. 

The memory used by models is presented for 32-bit (FP32) and 8-bit (compressed DRQ) data formats. The memory requirements of the DRQ and INT8 are very close. We have estimated the 32-bit memory of the literature models based on their number of parameters. This cannot be accurately used for the 8-bit memory as it is highly dependent on the quantization method and the model details. We should consider both the overall accuracy and the accuracy across individual classes for a fair comparison of the memory. The developed T model needs 328\,kB in 32-bit and only 139\,kB in 8-bit format with a good accuracy across all classes (F1\,$>$\,85\%).  

The accuracy of the compressed T model is lower than that of the compressed L model, while its memory requirement is 10x smaller. Although both models can be implemented on a commercial MCU, the T model can provide much faster run-time and smaller memory requirement. A key point is the medical importance of the accuracy advantage of the L model (up to 4 percentage points for the RBBB class) which can justify the deployment of a larger model.

\section{Conclusion}

In this paper, we presented a resource-efficient approach for the classification of heartbeat data from the MIT-BIH Arrhythmia Database. We proposed a multi-feature fusion strategy using time intervals and under-the-curve areas extracted from the ECG signal, demonstrating that this fusion significantly improves the accuracy of classification. Multiple Bi-LSTM based neural networks were implemented with varying sizes. The models were compressed using post-training quantization, including 8-bit integer and dynamic range quantization, and achieved state-of-the-art memory-efficient performance.

\end{document}